\documentclass{article} 
\usepackage{iclr2021_conference,times}
\usepackage{amssymb}
\usepackage{tabularx}
\usepackage{graphicx}
\usepackage{pgfplots} 
\usepackage{algorithm}
\usepackage{algpseudocode}
\usepackage{comment}
\usepackage{amsmath,amsfonts,bm}

\usepackage{hyperref}
\usepackage{url}
\usepackage{xcolor}
\usepackage{booktabs}
\hypersetup{
    colorlinks,
    linkcolor={red!50!black},
    citecolor={blue!50!black},
    urlcolor={blue!80!black}
}
\usepackage{listings}
\usepackage{subfiles} 

\renewcommand*{\thefootnote}{\fnsymbol{footnote}}

\lstdefinelanguage{Ini}
{
    basicstyle=\ttfamily\footnotesize,
    columns=fullflexible,
    morecomment=[s][\bfseries]{[c}{]},
    morecomment=[s][\bfseries]{[t}{]},
    commentstyle=\color{gray}\ttfamily,
    morekeywords={},
    otherkeywords={},
    keywordstyle={\color{green}\bfseries}
}

\title{Multi hash embeddings in spaCy}

\author{\parbox{\linewidth}{Lester James Miranda\footnotemark[1], Ákos Kádár\footnotemark[1], Adriane Boyd,   Sofie Van Landeghem, Anders Søgaard, Matthew Honnibal} \\
\\
ExplosionAI GmbH \\
}

\iclrfinalcopy 

\begin{document}

\maketitle
\footnotetext[1]{Equal contribution and corresponding authors: \texttt{\{lj,akos\}@explosion.ai} }

\renewcommand*{\thefootnote}{\arabic{footnote}}
\setcounter{footnote}{0}

\begin{abstract}
The distributed representation of symbols is one of the key technologies in
machine learning systems today, playing a pivotal role in modern natural
language processing. Traditional \emph{word embeddings} associate a separate
vector with each word. While this approach is simple and leads to good
performance, it requires a lot of memory for representing a large vocabulary. To
reduce the memory footprint, the default embedding layer in spaCy is a
\emph{hash embeddings} layer.  It is a stochastic approximation of traditional
embeddings that provides unique vectors for a large number of words without 
explicitly storing a separate vector for each of them. To be able to compute
meaningful representations for both known and unknown words, hash embeddings
represent each word as a summary of the normalized word form, subword
information and word shape.  Together, these features produce a
\emph{multi-embedding} of a word. In this technical report we lay out a bit of
history and introduce the embedding methods in spaCy in detail. Second, we
critically evaluate the hash embedding architecture with multi-embeddings on
Named Entity Recognition datasets from a variety of domains and languages. The
experiments validate most key design choices behind spaCy's embedders, but we
also uncover a few surprising results.
\end{abstract}

\section{Introduction}
The Python package spaCy\footnote{\url{https://spacy.io}} is a popular suite of
Natural Language Processing software, designed for production use-cases. It
provides a selection of well-tuned algorithms and models for common NLP tasks, 
along with well optimized data structures. The library also pays careful
attention to stability, usability and documentation.

Early versions of spaCy assumed that users would mostly use the default models
and architectures, and did not offer a fine-grained API for to customize and
control training. This changed with the release of v3, which introduced a new
configuration system and also came with a newly designed and documented machine
learning system, Thinc.\footnote{\url{https://thinc.ai}} Together these give
users full control of the modelling details.

Over the years spaCy has developed its own model architectures and default
hyperparameters, informed by practical considerations that are not only
motivated by scoring well on a single standard benchmark. In particular, spaCy
prioritizes run-time efficiency on CPU, the ability to run efficiently on long
documents, robustness to domain-shift, and the ability to fine-tune the model
after training, without access to the original training data.

In this technical report, we focus on one of the more unusual features of
spaCy’s default model architectures: its strategy for embedding lexical items.
We explain how this layer operates, provide experiments that show its
sensitivity to key hyperparameters, and compare it to a more standard embedding
architecture on a few datasets.  Our experiments focus on how the embedding
layer functions in the context of spaCy’s default Named Entity Recognition
architecture.

We these experiments aim to provide useful background information to users who
wish to customize, change or extend spaCy’s default architectures in their own
experiments. 

\section{Background}

\begin{figure}[t]
    \centering
    \includegraphics[scale=0.4]{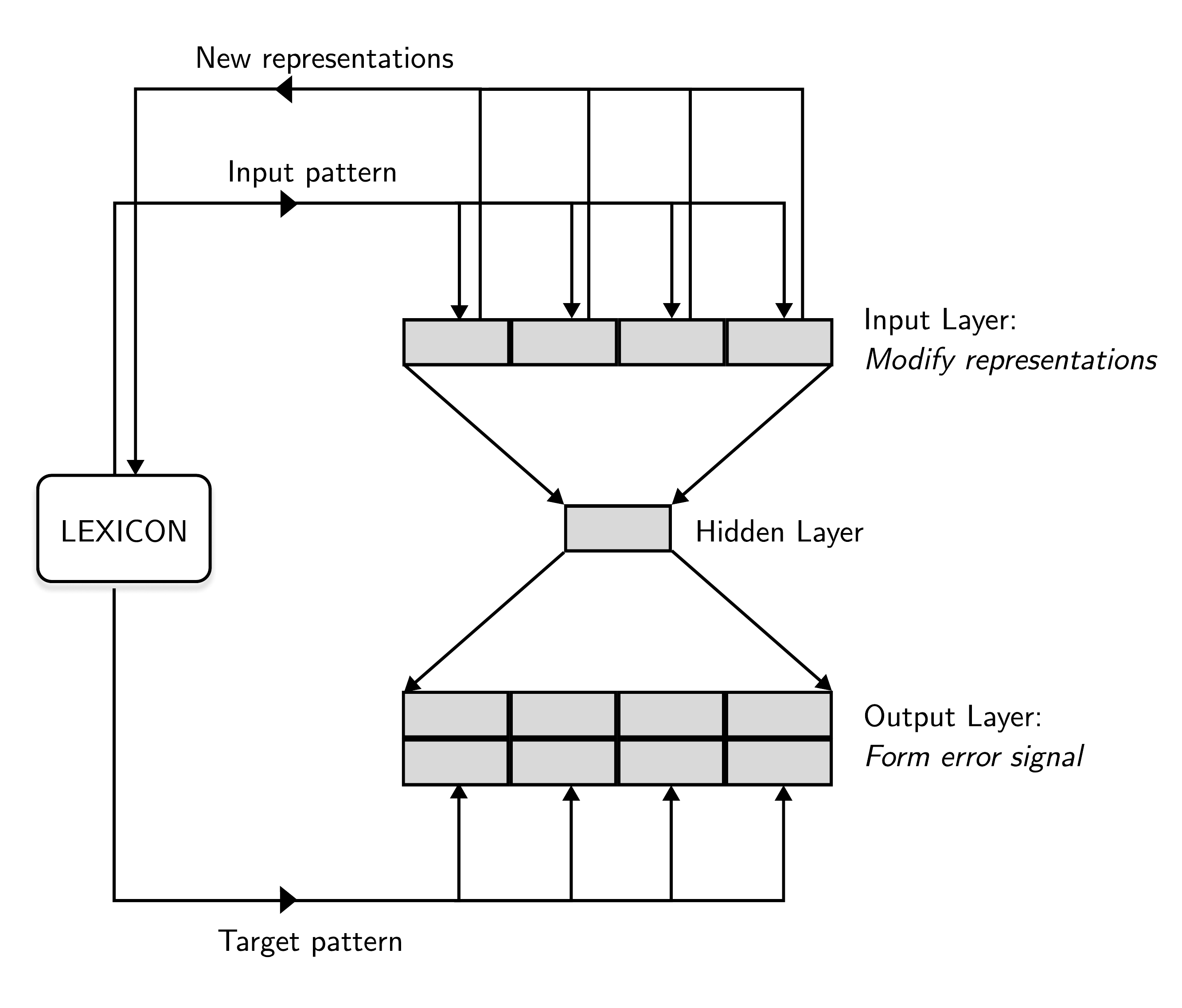}
    \caption{Illustration of the FGREP architecture. Redrawn from \cite{miikkulainen1991natural}}
    \label{fig:fgrep}
\end{figure}

Natural language processing (NLP) systems take text as input, broken down into a
list of words or subwords. Word embeddings have become a \emph{de facto}
standard for NLP, enabling us to represent compact tokens that can generalize
well.  These embeddings associate words with continuous vectors and are trained
such that functionally similar words have similar representations. The trained
embeddings encode useful syntactic and semantic information. While an
improvement over predecessors such as tf-idf, word embeddings are still  memory
intensive: each word in the vocabulary is typically stored as a 300-dimension
vector. As one of our main concerns in spaCy is efficiency, we use the hashing
trick to radically reduce the search space in a lookup table with little to no
degradation in accuracy.  To understand how spaCy's embedding system came to be,
let us go through a brief history of modern embeddings.

Real-valued representations of concepts have a long history. The way we think
about word embeddings today originates from the FGREP architecture of
\cite{miikkulainen1991natural}: a small neural network with a persistent public
global lexicon available in its input and output is learned end-to-end as shown
in Figure \ref{fig:fgrep}.  This global lexicon is a matrix where each row
vector represents a single word, and the distance (or angle) between these
vectors is small when words are similar and large when they are different. 

\citet{collobert2008unified} first popularized the idea of using neural network
models with pretrained word embeddings. They trained a convolutional
architecture to perform multiple linguistic tasks\textemdash named entity
recognition (NER), chunking, part-of-speech (POS) tagging, and semantic role
labeling\textemdash with a single neural network in a multi-task setting.
Crucially, they devised an efficient approach to pretrain word embeddings that
leverages large volumes of unlabeled text. To pretrain the embeddings, they 
constructed a simpler architecture than the one used for the downstream task,
and rather than maximizing the likelihood of the next word, as in
\cite{bengio2000neural}, they only trained the model to distinguish between real
samples from Wikipedia and \emph{contrastive} examples, i.e., sequences of words
sampled from Wikipedia where a random word replaces the middle word.  Their
follow-up paper, \emph{Natural Language Processing (Almost) from Scratch} 
\citep{collobert2011natural}, shows that cheap \emph{pretraining} of word
embeddings on large corpora closes the gap in downstream evaluation between
neural networks and NLP systems that rely on feature engineering. Roughly at the
same time, \cite{turian2010word} showed that adding word embeddings\textemdash
as well as more traditional Brown clusters \citep{brown1992class}\textemdash to
linear structured models of chunking and named entity recognition is a
straightforward way to boost performance.

The final key innovation that led to the widespread adoption of word embeddings
in the NLP community came when the pretraining phase was made cheaper. This was
accomplished by \cite{mikolov2013efficient} who removed the hidden layers and
non-linearities while relying entirely on linear models with contrastive losses.
To improve the coverage of word embeddings the popular library
fastText\footnote{\url{https://fasttext.cc}} computes word representations as
the sum of word and subword vectors
\citep{bojanowski2017enriching}.\footnote{Since then, various pretraining
objectives have been devised to train \emph{context-sensitive} word
representations by training a transformer encoder together with subword
embeddings. Some make use of masking, i.e., predicting missing words from
surrounding contexts \citep{devlin2019bert}, while others involve
self-supervision schemes like denoising corrupted documents
\citep{lewis2020bart}. These architectures are out-of-scope for this report, but
are available in spaCy under
\url{https://github.com/explosion/spacy-transformers}.}

The blueprint has not changed much since then. Today, the idea of pretraining
seems obvious and spaCy models are shipped with \emph{static vectors} pretrained
on large corpora. As we will show in the results for named entity recognition,
pretrained embeddings have a dramatic impact on the performance of modern NLP
systems. All models in spaCy can make use of pretrained embeddings, but also
learn word embeddings by backpropagating errors from the downstream tasks such
as named entity recognition or coreference resolution all the way down to the
word representations. 

Since we value efficiency, we use \emph{hash embeddings} when initializing
task-specific embeddings rather than storing separate representations for each
word. By applying the hashing trick
\citep{moody1988fast, weinberger2009feature}, we dramatically reduce the
vocabulary size to save memory \citep{tito2017hash}. Instead of performing a
simple lookup operation each word is hashed multiple times and the hashes are
modded into a vector table. Each word is then represented by a combination of
vectors.  As such hash embeddings represent a large number of words using a much
smaller number of vectors.  While spaCy models historically use standard static
vectors, in the future we are planning to release pretrained hash embeddings
trained with our floret
library\footnote{\url{https://github.com/explosion/floret/}} that combines
fastText with hash embeddings. 

In this report, we describe hash embeddings and compare their performance to
that of standard word embedding lookups. In addition, we fuse subword and word
shape information into the word representations rather than representing each
orthographic form with a vector. This process helps alleviate the issue of word 
forms being unseen during training and establish links between words that are
morphologically related.  We perform this comparison across multiple named
entity recognition benchmark datasets in multiple languages and domains.

\section{Embedding layers} 
Language processing deals with inputs that are made up of discrete symbols.  A
word embedding layer $\varepsilon$  is a mapping from entries from a dictionary
to vectors.  Task-specific embedding functions are often learned in an
end-to-end fashion, in which case $\varepsilon$ adapts the representation of
words to the downstream task at hand while preserving relevant linguistic 
regularities. General-purpose embeddings, on  the other hand, are trained in a
fully self-supervised fashion, typically to predict words from contexts  or with
a contrastive loss. The simplest representation of input tokens is to encode the
\emph{identity} of each symbol, and map  each $s \in \Sigma$ in our vocabulary
$\Sigma$ to a binary \emph{one-hot} vector $\mathbf{b} \in \{0, 1\}^{|\Sigma|}$.
Here $\sum_j \mathbf{b}_j = 1$, meaning that only one element of the vector is
set to 1, and all others are 0. Such a representation disregards any information
about how symbols are used, and its dimensionality grows  linearly with the
number of input symbols.

\begin{figure}[t]
    \centering
    \includegraphics[width=0.95\textwidth]{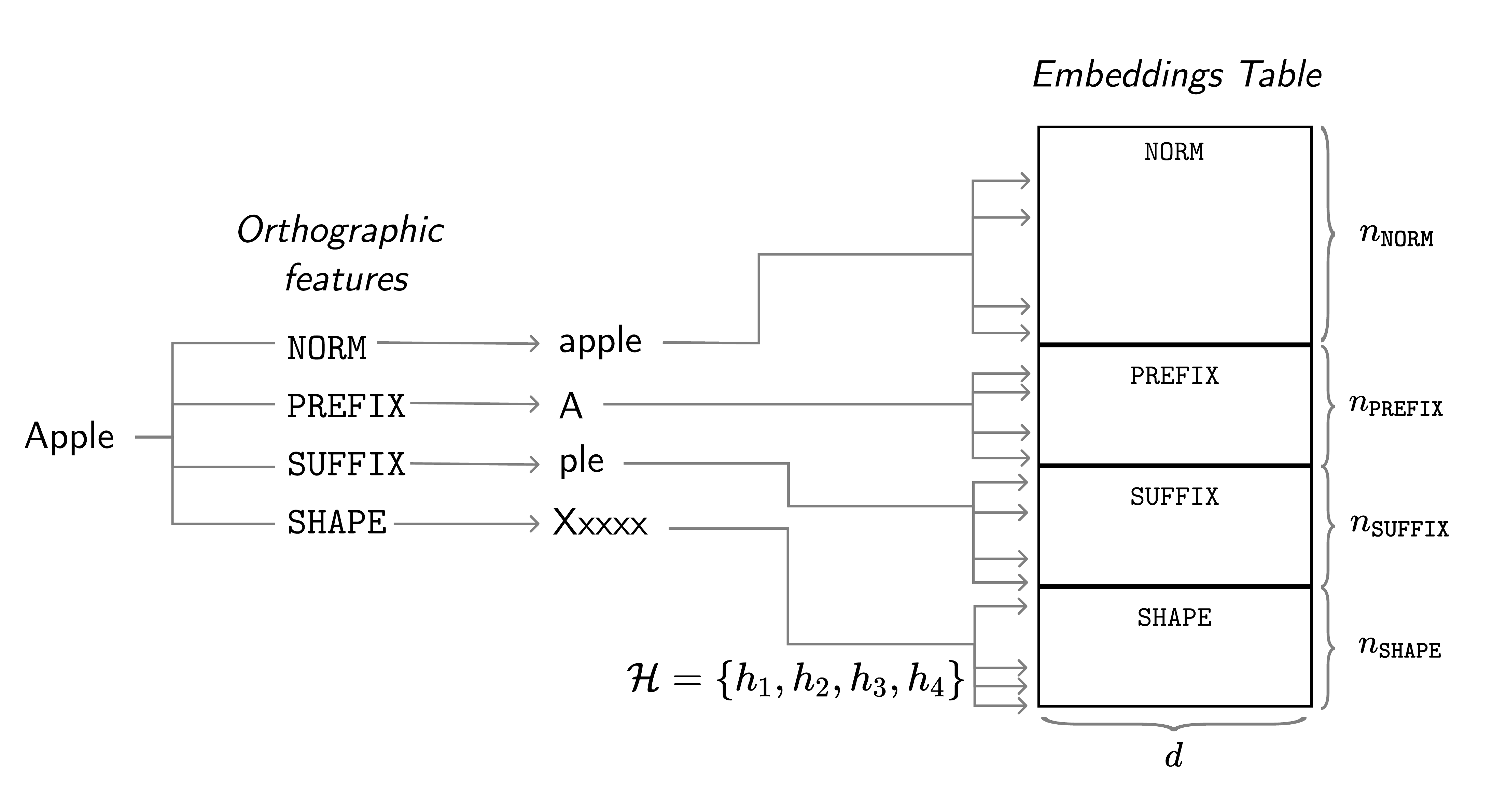}
    \caption{
        Diagram of the \texttt{MultiHashEmbed} algorithm. First, from the
        orthographic form of ``Apple'' various features are extracted. Then each
        feature is hashed four times and modded to their feature-specific
        tables. The four vectors per table summed up and the resulting pooled
        feature vectors are fed to a Maxout layer to produce the final
        embedding.
        }
    \label{fig:mhe}
\end{figure}

\begin{algorithm}
\caption{One-hot encoding}
\label{alg:to_one_hot}
\begin{algorithmic}[1]
\Require{$s, \Sigma$ \Comment{Symbol, vocabulary}}
\Ensure{$\mathbf{b} \in \{0, 1\}^{\vert \Sigma \vert}$ \Comment{One-hot vector}}

\State {$N$ $\gets$ {$\vert \Sigma \vert$}}
\State {$i \gets index(\Sigma, s)$}
\State {$\mathbf{b} \gets \{\vec{\scriptstyle 0}\}$, $\vert \{\vec{\scriptstyle 0}\} \vert = N$} 
\State {$\mathbf{b}_i \gets 1$}
\end{algorithmic}
\end{algorithm}

It has become the \emph{de facto}~standard in NLP to  define a fixed width 
embedding matrix  $\mathbf{E}_{\ell} \in \mathbb{R}^{|\Sigma| \times d}$ where
width $d$ is a hyperparameter, and embed each symbol as $\mathbf{E}_{\ell}^T
\mathbf{b}$.\footnote{We use the subscript $\ell$ for the ``lookup'' operation
performed when indexing into $\mathbf{E}$} Rather than using a dot-product, this
process is implemented as a lookup operation using a mapping $\texttt{map}:
\Sigma \mapsto \mathbb{N}$:

\begin{equation}
    \varepsilon(s) = \mathbf{E}_{\ell}[\texttt{map}(s)]
\end{equation}

The embedding layers constitute  the input layer, and during training, the
gradients from the loss function are backpropagated all the way down to
$\mathbf{E}_{\ell}$, through all upstream layers. This process adapts
$\mathbf{E}_{\ell}$ to provide representations of symbols that are useful to
solve the downstream task.

\begin{algorithm}
\caption{Create embedding table}
\label{alg:make_table}
\begin{algorithmic}[1]
\Require{$d, \texttt{map}$ \Comment{Dimensions, vocabulary}}

\State {$\mathbf{E}_{\vert \Sigma \vert \times d} \sim U(0,1)$} \Comment{Uniform random distribution}
\Procedure{embed}{$s$, $\texttt{map}$}
    \State {$i \gets \texttt{map}(s)$}
    \State \Return {$\mathbf{E}_{i}$}
\EndProcedure
\end{algorithmic}
\end{algorithm}

To keep the method tractable, we use the common practice of choosing a threshold
to determine the minimum frequency of a token that is worth embedding.  Another
popular criterion is to fix the size $k$ of the embedding table  
$\mathbf{E}_{\ell}$ and only learn vectors for the top-$k$ most frequent
symbols.  All symbols that are left out of $\mathbf{E}_{\ell}$ get mapped to the
special \texttt{UNK} symbol, and its corresponding vector is used to represent
all unseen words.

\subsection{Hash Embedding layer}

Rather than storing a separate vector for each symbol, hash embeddings apply the
hashing trick in order to reduce the memory footprint. This method is inspired
by Bloom filters \citep{bloom1970space}, a simple probabilistic data structure
to solve the membership problem, i.e., to answer the question of whether we have
seen an element $s$ before.  Bloom filters only have two operations:
\emph{inserting} an element and \emph{testing} whether an element has already
been inserted.  They represent a large set $S$ by a compact bit vector
$\mathbf{b} \in \{0, 1\}^n$, where we choose $n$ to be small $|S| \gg n$. 
Before an element is inserted, it is hashed by $k$ number of uniform hash
functions: given $n$ buckets, each function maps each symbol $s \in S$ to any
bucket with probability $\frac{1}{n}$. The hash functions are also from a 
$k$-independent family to guarantee that they are not correlated.  Insertion of
$s$ is performed by hashing it $k$ times and setting the corresponding bits in
$\mathbf{b}$ to 1. Testing whether $s$ is in the filter is done by performing
the hashing and looking up the corresponding values. If all values are 1 then
the element \emph{may exist} in the filter otherwise it \emph{definitely does
not exist}.

Similarly, hash embeddings are parametrized by the number of rows $n$, the width
$d$ (like usual embedding tables), and by the number of mutually independent
hash functions $\mathcal{H} = \{h_1 \ldots h_k\}$ as shown in Figure
\ref{fig:mhe}. Each incoming symbol $s$ is hashed $k$ times and modded into the
embedding table $\mathbf{E}_{h} \in \mathbb{R}^{n\times d}$.\footnote{We use the
subscript $h$ for the "hash" operation performed when indexing into
$\mathbf{E}_{h}$} Given a list of hash functions, the embedding of the symbol
$s$ is:

\begin{equation}
    \varepsilon(s) = \sum_{h \in \mathcal{H}} \mathbf{E}_h[h(s)\,\mathbin{\%} \, n] 
\end{equation}

\begin{algorithm}
\caption{An algorithm for hash embed}
\label{alg:hashembed}
\begin{algorithmic}[1]
\Require{$\mathbf{E}, s, \mathcal{H}$, n \Comment{Embedding table, symbol, hash functions, number of rows}}
\Ensure{$\mathbf{S}$ \Comment{Pooled feature vector}}

\State {$\mathbf{S} \gets  \{\vec{\scriptstyle 0}\}$}
\State {$N$ $\gets$ {$\vert\mathcal{H}\vert$}}
\For{$k \gets 1$ to $N$}
    \State $i \gets \mathcal{H}_{k}(s) \mathbin{\%} n $ 
    \State $\mathbf{S} \gets \mathbf{S} + \mathbf{E}_i$ \Comment{Vector sum of $i$-th row}
\EndFor
\end{algorithmic}
\end{algorithm}

Recall that in the regular embedding layer we had a one-hot indicator vector
$\mathbf{b}$.  In the case of hash embeddings, the indicator is multi-hot where
$\sum_j \mathbf{b}_j = k$. Using the multi-hot indicator vector in matrix form,
the multiple selection can be written analogously to the usual embedding layer
$\mathbf{E}_h^T\mathbf{b}$. Just like traditional embeddings, hash embeddings
produce a single vector for each $s$. However, the main advantage of hash
embeddings is that we do not need to store a separate vector for each symbol.
Since each symbol is represented as a sum of $k$ vectors, a large number of
\emph{signatures} can be generated as a combination of a small number of
vectors.

\subsection{Collisions}

Hash embeddings just like Bloom filters are prone to collisions. Let's assume
that due to the uniformity of the hash functions, each symbol $s \in \Sigma$ is
mapped to any row with probability $\frac{1}{n}$.  Conversely, the probability
for a specific row \emph{not} being chosen by a single hash function is $1 -
\frac{1}{n}$.  The collision probability $p_c$ of a hash function with range $0$
to $n$ over a vocabulary of size $|\Sigma|$ is $1 - (1 - \frac{1}{n})^{|\Sigma|
- 1}$. If we have 50,000 distinct words in our corpus, a table of 5,000 rows,
and a single hash function, then the probability that some word will be mapped
onto the same vector as another is 0.99995. The expected number of tokens that
will collide is $p_c \times |\Sigma|$, which in this case is 49,663. We confirm
that using a small table with a single hash function can cause most tokens to
collide.  However, by using $k$ independent hash functions with range $\{0,
\ldots, n\}$, we approximate a hash function with a greater upper-bound $n^k$.
In our case, we set $k=4$, yielding $p_c \approx 5 \times 10^{-12}$. This means
that in theory, we can use significantly fewer parameters than we have symbols,
with a negligible collision risk.

\subsection{Multi-embeddings with orthographic features}

When talking about word embeddings, it is not always clear which symbols are
embedded by NLP architectures.  Sometimes, a \emph{word} can mean its raw
orthographic form, but at other times, it can also be its lowercase or other
normalized form. The \texttt{Tokenizer} component in spaCy extracts various
features from a token's orthographic representation for downstream 
embedding.\footnote{\url{https://spacy.io/api/token\#attributes}} We embed the
following features:

\begin{itemize}
    \item\texttt{NORM}: Lowercased token with additional normalizations
    regarding currency symbols, punctuation and alternate spelling.
    \item\texttt{PREFIX}: First character.
    \item\texttt{SUFFIX}: Last three characters.
    \item\texttt{SHAPE}: Alphabetic characters are replaced by \texttt{x} or
    \texttt{X} based on casing, numeric characters are replaced by \texttt{d}
    and sequences of the same character are truncated after length 4.
\end{itemize}

We denote the embedding of \texttt{NORM}, \texttt{PREFIX}, \texttt{SUFFIX} and
\texttt{SHAPE} features by $\mathbf{e}^{\text{norm}}$, 
$\mathbf{e}^{\text{prefix}}$, $\mathbf{e}^{\text{suffix}}$ and
$\mathbf{e}^{\text{shape}}$, respectively.  The multi-embedding layers take a
global \emph{width} parameter, setting the dimension $d$ for each embedding.
The embeddings are then concatenated $[\mathbf{e}^{\text{norm}};
\mathbf{e}^{\text{prefix}}; \mathbf{e}^{\text{suffix}};
\mathbf{e}^{\text{shape}}]$, producing a single embedding of size $4d$. The
concatenated vectors are then projected down to size $d$ by a small neural
network $\mathbf{e} = \Psi([\mathbf{e}^{\text{norm}};
\mathbf{e}^{\text{prefix}}; \mathbf{e}^{\text{suffix}};
\mathbf{e}^{\text{shape}}])$. In spaCy, we use a Maxout-layer
\citep{goodfellow2013maxout} as represented by the function $\Psi:
\mathbb{R}^{4d} \mapsto \mathbb{R}^{d}$:

\begin{equation}
    \Psi(\mathbf{x}) = \max(\mathbf{W}_1^T\mathbf{x} + \mathbf{b}_1, \mathbf{W}_2^T\mathbf{x} + \mathbf{b}_2, \mathbf{W}_3^T\mathbf{x} + \mathbf{b}_3) 
\end{equation}

Here, the matrices $\mathbf{W}_1, \mathbf{W}_2, \mathbf{W}_3$ of size $4d \times
d$ and bias terms $\mathbf{b}_1, \mathbf{b}_2, \mathbf{b}_3$ of size $d$ are the
learnable parameters of the network.  The maxout layer is implemented as a
component-wise maximum over multiple linear layers. In spaCy, we use three
pieces for the multi-embedding layers. Our standard embedding layer,  
\texttt{MultiHashEmbed}, combines the multi-embedding process described here
with hash embeddings.  For this technical report, we additionally implemented
\texttt{MultiEmbed}, which is identical to \texttt{MultiHashEmbed} except that
it does not use the hashing trick.

\section{Experimental Setup}

The main goal of our experiments is to benchmark our hash embedding
implementation, \texttt{MultiHashEmbed}, on different settings and scenarios
against traditional word embeddings.  This section outlines the datasets we used
as well as our model architecture. We tested on a variety of named entity
recognition datasets from multiple domains. For word embeddings we used the
vectors distributed with spaCy 3.4.3 (large)
models.\footnote{Section~\ref{sec:spacyfasttext} in the Appendix compares these
embeddings to fastText vectors.}

\subsection{Named Entity Recognition Datasets}
We selected datasets from a variety of domains and languages and that are of
moderate size, which represent realistic use cases for spaCy. 

\paragraph{CoNLL 2002 \citep{tjong-kim-sang-2002-introduction}} 
Standard benchmark containing news articles. For Spanish the corpus contains
NewsWire articles from May 2000.  The Dutch data come from the Belgian newspaper
\emph{De Morgen} from the months of June to September 2000.  We included this
dataset as it is one of the longest standing standard benchmarks in the field.
Entities include: \textit{location, person, organization, misc}. 

\paragraph{WNUT 2017 \citep{derczynski-etal-2017-results}} Standard NER
benchmark based on English social media data with the same training data as the
Twitter NER corpus \citep{derczynski2016broad} annotated with tweets from 2010.
This dataset is especially challenging, because the test set contains entities
not seen during training. Entities include: \textit{location, person, group,
corporation, creative-work, product}.

\paragraph{AnEM \citep{ohta-etal-2012-open}} NER dataset focused on anatomical
entity extraction. The articles were sourced from the PubMed Database of
publication abstracts and the PubMed Central (PMC) Open Access subset of
full-text publications. This dataset is interesting due to its specialized
domain, where pretrained general-purpose word embeddings are not expected to
provide the same performance boost as for news texts, for example.  Entities
include: \textit{cell, organism substance, pathological formation, multi-tissue
structure, organism subdivision, organ, cellular component, anatomical system,
tissue, developing anatomical structure, immaterial anatomical entity}.

\paragraph{Dutch Archaeology \citep{brandsen-etal-2020-creating}} Contains
excavation reports and related documents collected since the 1980s. The texts
were gathered by the Digital Archiving and Networked Services (DANS) in the
Netherlands between 2000 and 2020. This is also a dataset from a specialized
domain and entities include: \textit{time period, location, context, artefact,
material, species}.

\paragraph{OntoNotes 5.0 \citep{weischedel2013ontonotes}} A standard NER
benchmark that contains text from various genres (e.g.,  news, phone
conversations, websites, etc.). We included this dataset for its larger
vocabulary size. Entities include: \textit{date, geopolitcal entity, ordinal,
organization, quantity, location, cardinal, person, nationalities or religious
or political groups, facility, time, event, money, work of art, law, percent,
product, language}.

\subsection{Dataset processing details}

\paragraph{Preprocessing} The Dutch CoNLL 2002 data contains document-level
segmentation, which we use in our experiments. This is the only document-level
task and for all other datasets we rely on the provided sentence segmentation.
The AnEM dataset only contains training and test splits. We define a random
split of 80\% of the training set for training and 20\% for development and use
the original test set for testing. The Dutch Archaeology dataset does not
provide canonical splits, so we create a random split of 80\% for training, 10\%
for development and 10\% for testing. Lastly, we modified the English version of
OntoNotes 5.0 by adding raw texts and normalizing punctuation.
Table~\ref{tab:dataset_statistics} in the Appendix gives an overview of 
characteristics of the training sets. 

\paragraph{Unseen evaluation} To get a better sense of real-world expected 
performance differences between \texttt{MultiEmbed} and \texttt{MultiHashEmbed},
we include a separate evaluation for \emph{unseen} test entities. Since entity
IDs are not available for all datasets, we consider each span as an unseen
entity as long as it does not appear verbatim in the training set. Unseen
entities are evaluated by ignoring all known entity spans during evaluation.

\subsection{Model architecture and training details}

\paragraph{Named Entity Recognizer architecture} The named entity recognizer
model in spaCy is transition-based \citep{lample2016neural}, manipulating an
input buffer of tokens and a stack of partially constructed structures. It
relies on the BILUO sequence encoding scheme to determine whether tokens are at
the beginning (Begin), in the middle (In) or at the end (Last) of an entity,
make up single-token entities (Unit) or are not part of an entity at all (Out).
The named entity recognizer makes these decisions based on the current word as 
well as the words from the current entity in the stack if one exists: its first
word, its last word, and three additional context tokens. 

Given these state representations, a single-layer Maxout network
\citep{goodfellow2013maxout}  computes a state vector and another feed-forward
network with a Softmax activation computes the action probabilities from that
state vector.  To train the model, a dynamic oracle \citep{goldberg2012dynamic}
is used with an imitation learning objective that scores how many correct
entities are produced while avoiding the prediction of incorrect entities. 
Overall, this algorithm is well suited to named entity recognition problems for
languages similar to English, where named entities generally have distinct
boundary tokens.

\paragraph{Training details} For all experiments we use the same model
architecture and only vary the embeddings. We refer to hash embeddings with
multiple orthographic features as 
\texttt{MultiHashEmbed}.\footnote{\url{https://spacy.io/api/architectures\#MultiHashEmbed}}
For comparison, we implemented \texttt{MultiEmbed} that is the same as
\texttt{MultiHashEmbed}, but using regular lookup instead of the hashing trick.
Both embedding layers use the \texttt{NORM}, \texttt{PREFIX}, \texttt{SUFFIX}
and \texttt{SHAPE} features; for \texttt{MultiHashEmbed} we use 5000, 2500, 2500
and 2500 rows for each table respectively and for \texttt{MultiEmbed} we use the
minimum frequency criterion of 10.\footnote{For results on the effect of minimum
frequency see Section~\ref{sec:minfreq} in the Appendix.} We use pretrained
static vectors from the large (\texttt{lg}) models distributed with spaCy v3.4.

The embeddings for the orthographic features are concatenated and fed through a
single Maxout layer with three pieces. The token vectors from the embedding
layer are passed to an 8-layer convolutional encoder with residual
connections\textemdash all of which have a window size of 3, width of 96, and
Maxout activations. We also use layer normalization \citep{ba2016layer} after
each layer. 

During training we use dropout \citep{srivastava2014dropout} with probability
0.1.  The training is run for a maximum of 20000 training steps with a patience
of 1600 steps and validation frequency of 200. We use a batch size of 1000
words. For the optimizer we use AdamW \citep{loshchilov2018decoupled} with
$\beta_1=0.9$ and $\beta_2=0.999$ and apply weight decay with coefficient
$0.01$. The starting learning rate is set to $0.001$ and gradient clipping is
applied with norm $1.0$.\footnote{The spaCy config files are shown in
Section~\ref{sec:configuration} in the Appendix.}

\section{Results}
\label{results}

This section reports the results for different benchmarking scenarios for
\texttt{MultiHashEmbed}. For all experiments, we report the average F1-score
across three random seeds. We included the full results in tables in the
Appendix.

\subsection{Comparing \texttt{MultiEmbed} and \texttt{MultiHashEmbed} embedding strategies}
\label{sec:mhe_v_me}

\begin{figure}[t]
    \centering
    \includegraphics[width=\textwidth]{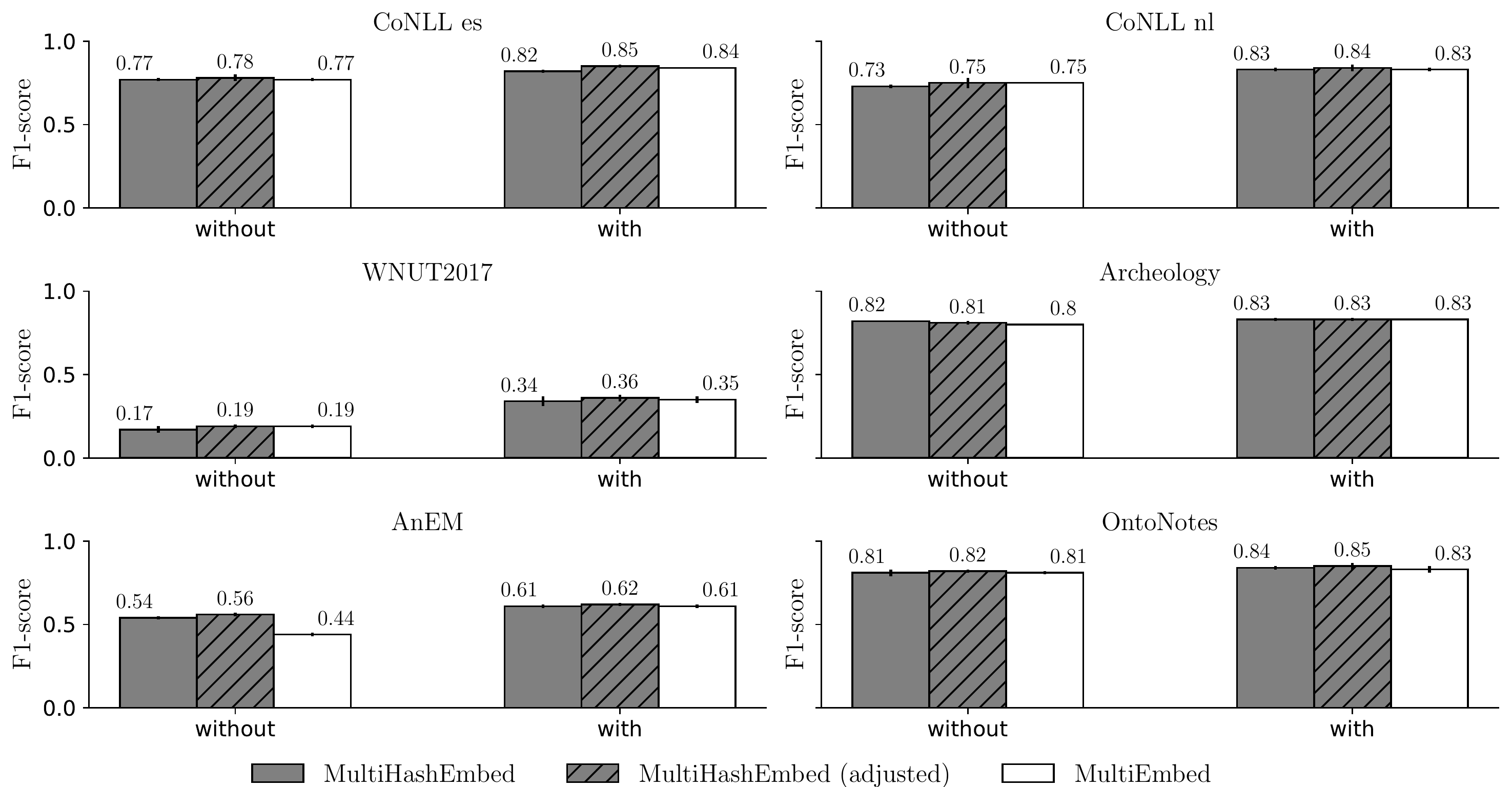}
    \caption{
        Comparing the \texttt{MultiHashEmbed} and \texttt{MultiEmbed} embedding
        strategies with and without pretrained embeddings across a variety of
        NER datasets. Evaluated on the test set.
        }
    \label{fig:mhe_v_me}
\end{figure}

\begin{figure}[t]
    \centering
    \includegraphics[width=\textwidth]{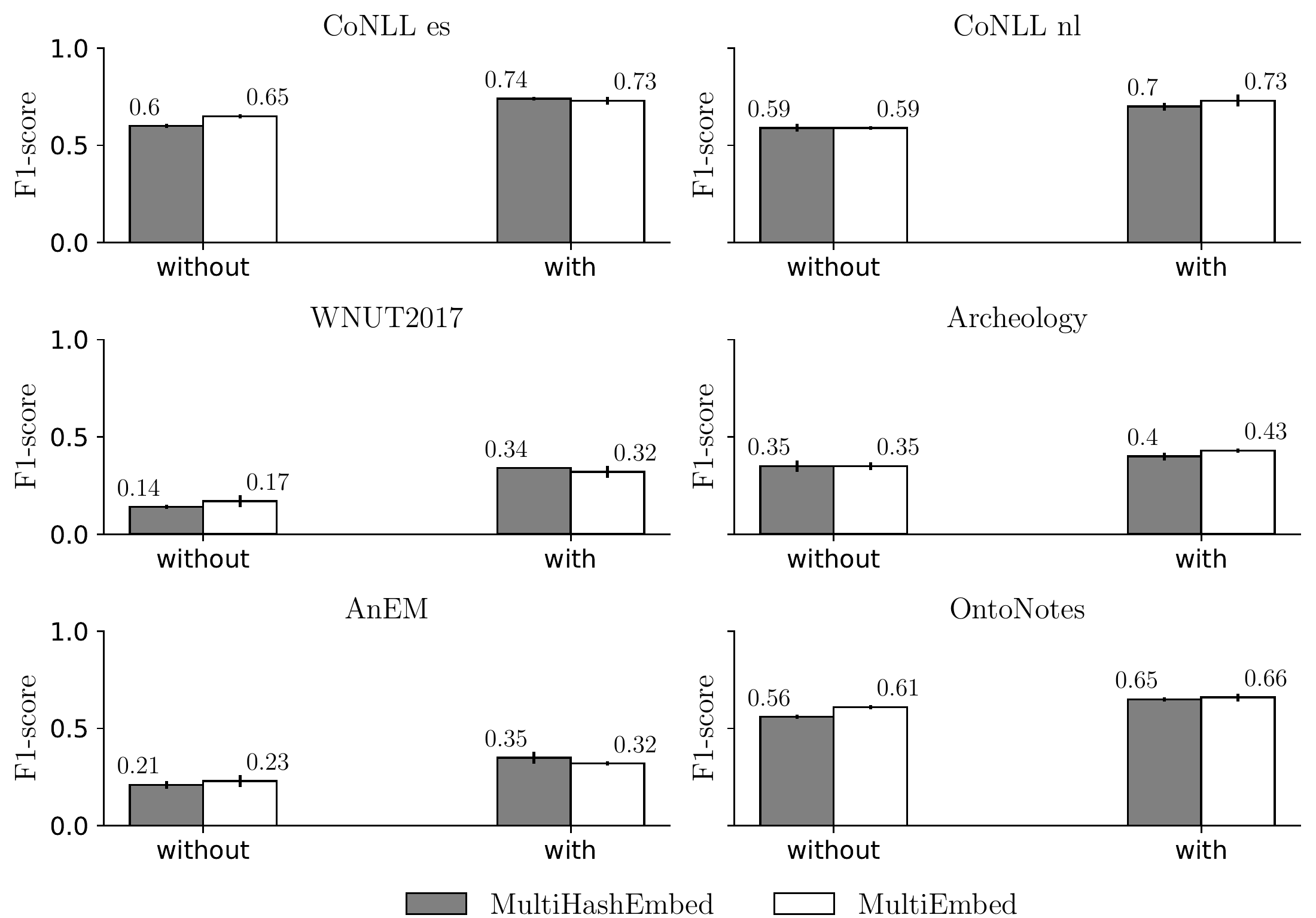}
    \caption{
        Comparing the \texttt{MultiHashEmbed} and \texttt{MultiEmbed} embedding
        strategies with and without pretrained embeddings across a variety of
        NER datasets. Evaluated on unseen test entities.
        }
    \label{fig:mhe_v_me_unseen}
\end{figure}

We compare \texttt{MultiEmbed} and \texttt{MultiHashEmbed} with and without the
use of pretrained embeddings. To level the playing field we also added an
\emph{adjusted} setup for \texttt{MultiHashEmbed}, where we set the number of
rows in the lookup tables equal to that of \texttt{MultiEmbed}. This is because
\texttt{MultiEmbed} peeks at the data and adjusts its size based on the
frequencies of symbols in the data and applying the minimum frequency filtering,
whereas the defaults of \texttt{MultiHashEmbed} are designed to work well
without knowing the details of the dataset.

Figure~\ref{fig:mhe_v_me} shows the results.\footnote{The complete results are
presented in Tables~\ref{tab:mhe_v_me_default_spacy_test},
\ref{tab:mhe_v_me_adjusted_rows_spacy_test},
\ref{tab:mhe_v_me_default_null_test} and \ref{tab:mhe_v_me_defaults_null_test}
in the Appendix.} When comparing the upper row with the lower row we find a
consistent benefit from using pretrained embeddings across all datasets. Based
on these results \textbf{we recommend to use pretrained embeddings when
possible}. Notably, however, the benefit of pretraining is the least pronounced
for the Archeology and OntoNotes datasets. This is probably due to OntoNotes
being a much larger dataset than the others and Archeology having a comparable
vocabulary size to the CoNLL datasets while having 350k tokens in the training
set compared to the 200k and 260k of the Dutch and Spanish CoNLL respectively. 

Comparing the hash embeddings to the traditional embeddings we find strong
evidence (Figure \ref{fig:mhe_v_me}) that our hash embeddings with default
number of rows result in the same performance compared to traditional
embeddings. When peeking into the dataset and adjusting the rows we can get only
a slight increase in performance across the board. This is interesting, because
on most datasets the number of vectors in the \texttt{MultiHashEmbed} tables end
up being \emph{more} than \texttt{MultiEmbed} when using the default minimum
frequency criterion of 10. However, on OntoNotes the hash embeddings have half
the number of rows for \texttt{NORM} as \texttt{MultiEmbed} and the performance
is still equivalent. This shows that the defaults work in general, but also 
points out the memory savings of the  hashing trick in practice.

Lastly, the results on unseen entities as shown in Figure
\ref{fig:mhe_v_me_unseen} suggest that \texttt{MultiEmbed} and 
\texttt{MultiHashEmbed} also perform similarly on unseen entities.\footnote{The
complete results on unseen entities are presented in
Tables~\ref{tab:mhe_v_me_unseen_default_null_test} and
\ref{tab:mhe_v_me_unseen_default_spacy_test} in the Appendix.} However, it's
worth noting that systems incur a significant drop in performance when evaluated
on entities not seen  during training even when using pretrained
embeddings.\footnote{The results in WNUT 2017 are almost the same, because
almost all entities are unseen.}

We conclude that \texttt{MultiHashEmbed} performs head to head with
\texttt{MultiEmbed}, highlighting the validity of the use of hash embeddings. We
also found that having more or fewer rows as \texttt{MultiEmbed} does not change
the conclusion showing robustness to this parameter and also pointing out the
memory benefits of the hashing trick.

\subsection{Number of Rows}

\begin{table}[t]
    \centering
    \begin{tabular}{cccccc}
        \toprule
         & \multicolumn{4}{c}{Number of rows in \texttt{MultiHashEmbed}} \\
                        &   Default & $=$ \texttt{MultiEmbed} & $=20\%$ \texttt{MultiEmbed} & $=10\%$ \texttt{MultiEmbed}  \\
         \midrule
        CoNLL es &     0.77$\pm$0.00      &  0.79$\pm$0.01 & 0.78$\pm$0.02  & 0.78$\pm$0.01  \\
        Archeology  &  0.83$\pm$0.01  & 0.83$\pm$0.01  & 0.82$\pm$0.02  & 0.80$\pm$0.01  \\
        \bottomrule
    \end{tabular}
    \caption{Results reported are F1-scores on the dev sets. Rows are adjusted based on the lookup table of \texttt{MultiEmbed}. Without pretrained embeddings.}
    \label{tab:mhe_rows_null}
\end{table}

The main motivation behind hash embeddings is to use a small amount of vectors
and still achieve good performance.  We have already seen that on OntoNotes:
\texttt{MultiHashEmbed} performs comparably to \texttt{MultiEmbed} even if it
only has half of the number of vectors for NORM.

Here we go one step further and test \texttt{MultiHashEmbed}, when using only
20\% or 10\% of the number of vectors available to \texttt{MultiEmbed}.
Table~\ref{tab:mhe_rows_null} shows the results on the development sets of the 
Spanish CoNLL and Dutch Archeology datasets without using pretrained vectors. On
the Spanish CoNLL dataset we find no degradation of performance even when using
as little as 10\% of the number of rows as \texttt{MultiEmbed}. On the Dutch
Archeology dataset using 20\% of the rows available to \texttt{MultiEmbed}
causes a 1\% drop in performance, while using 10\% leads to a 3\% drop. The
results show that \textbf{hash embeddings can achieve comparable performance to
traditional embeddings with significantly less parameters}. 

\subsection{Orthographic Features}
\label{sec:orth}

\begin{table}[t]
    \centering
    \begin{tabular}{cccccccc}
        \toprule
         \multicolumn{5}{c}{Feature Combination} & \multicolumn{3}{c}{Relative Error (F1-score)}\\
         ORTH & NORM & PREFIX & SUFFIX & SHAPE & All & Seen & Unseen \\
         \midrule
          & $\checkmark$ & $\checkmark$ & $\checkmark$ & $\checkmark$ & - & - & -  \\
         & $\checkmark$ & $\checkmark$ & $\checkmark$ &  & +17\% & +0\% &  +15\%  \\
         & $\checkmark$ & $\checkmark$ & & & +30\% & +80\% & +26\% \\
         & $\checkmark$ & & & & +47\% & +100\% & +68\% \\
         $\checkmark$ & & & & & +50\% & +160\% & +62\% \\
         \bottomrule
    \end{tabular}
    \caption{Relative error increase on \texttt{MultiHashEmbed} embedding given various combinations of orthographic features for the CoNLL Dutch dataset. Without pretrained embeddings.}
    \label{tab:mhe_feature_ablation_null_conll-nl-diff}
\end{table}

\begin{table}[t]
    \centering
    \begin{tabular}{cccccccc}
        \toprule
         \multicolumn{5}{c}{Feature Combination} & \multicolumn{3}{c}{Relative Error (F1-score)}\\
         ORTH & NORM & PREFIX & SUFFIX & SHAPE & All & Seen & Unseen \\
         \midrule
          & $\checkmark$ & $\checkmark$ & $\checkmark$ & $\checkmark$ & - & - & -  \\
         & $\checkmark$ & $\checkmark$ & $\checkmark$ &  & +4\% & -31\% &  +3\%  \\
         & $\checkmark$ & $\checkmark$ & & & +10\% & -46\% & +5\% \\
         & $\checkmark$ & & & & +10\% & -35\% & +12\% \\
         $\checkmark$ & & & & & +20\% & -35\% & +16\% \\
         \bottomrule
    \end{tabular}
    \caption{Relative error increase on \texttt{MultiHashEmbed} embedding given various combinations of orthographic features for the AnEM dataset. Without pretrained embeddings.}
    \label{tab:mhe_feature_ablation_null_anem-diff}
\end{table}

After comparing hash embeddings with traditional embeddings, we turn to
evaluating the contribution of the orthographic features. We start with spaCy's
default \texttt{NORM}, \texttt{PREFIX}, \texttt{SUFFIX} and \texttt{SHAPE}
features, then gradually remove them one-by-one while measuring their effect on
performance. We also included an \texttt{ORTH}-only configuration, which
represents the most common method outside of spaCy. Tables 
\ref{tab:mhe_feature_ablation_null_conll-nl-diff} and
\ref{tab:mhe_feature_ablation_null_anem-diff} report the relative error increase
in the F1-score for Dutch CoNLL 2002 and AnEM.  We used these two datasets
because CoNLL is a standard benchmark representing a common choice to tune
default parameters and architectures. In contrast, AnEM is a smaller dataset
with a specialized domain.

Table~\ref{tab:mhe_feature_ablation_null_conll-nl-diff} reports the results for
CoNLL Dutch, which are in line with our expectations: removing any of the
features degrades performance and \texttt{ORTH} performs the worst overall. We
do find the same pattern for the AnEM dataset in 
Table~\ref{tab:mhe_feature_ablation_null_anem-diff} but only if we consider the
global F1 score. However, when engaging with a more fine-grained analysis we see
that the error decreased on seen entities, but increased on unseen entities.
This highlights the need for more detailed evaluation processes to find emergent
differences between systems even when the standard evaluation seems to support
one's hypotheses and intuitions.\footnote{We will release a utility for 
evaluating on seen / unseen entities in the future.} Overall, our results show
that \textbf{subword and word shape features can be a cheap and effective way to
improve performance}.

\subsection{Number of hash functions}
\label{sec:hash_func}

\begin{figure}[t]
    \centering
    \includegraphics[width=\textwidth]{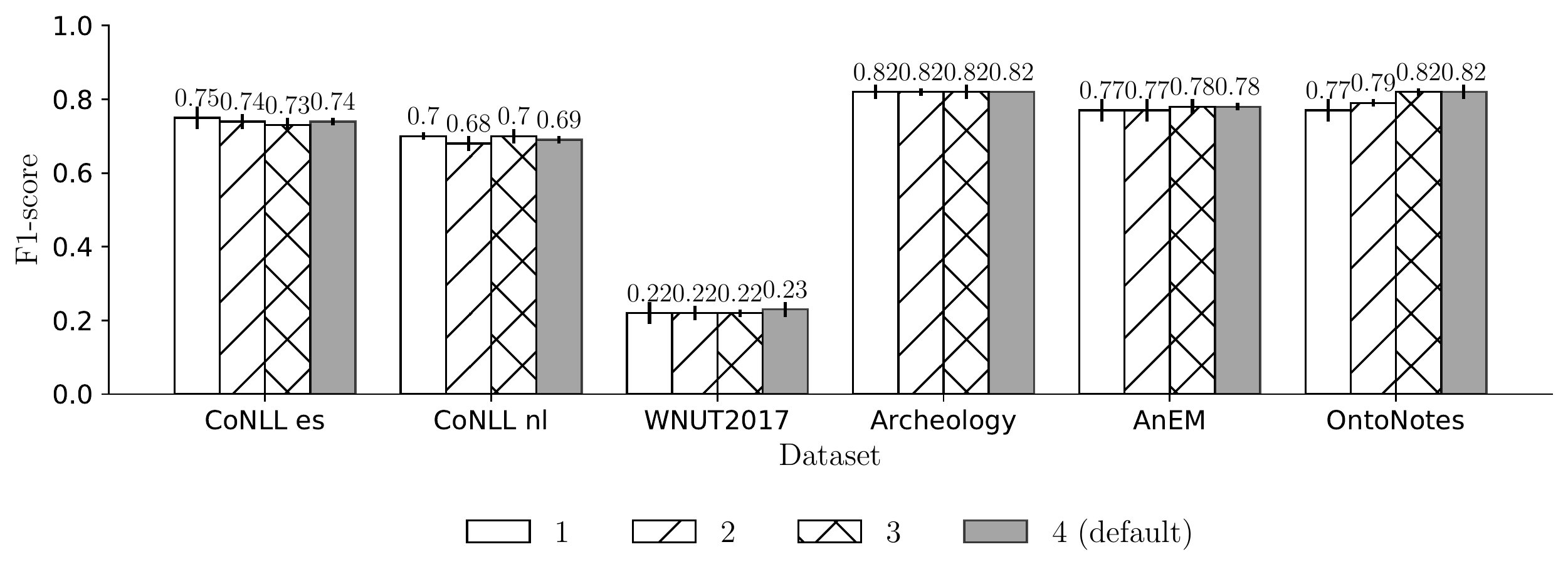}
    \caption{
        Performance of \texttt{MultiHashEmbed} with different number of hash
        functions. Evaluated on the development set.
        }
    \label{fig:collision}
\end{figure}

Other than the number of rows in the \texttt{MultiHashEmbed} tables the number
of independent hash functions can also be used to control the capacity of the
embedding layer. In spaCy this is currently fixed to four, but to critically 
evaluate this choice we tried the model with one, two, three and four hash
functions. The results in Figure~\ref{fig:collision} indicate that the
performance does not vary too much on most datasets.\footnote{Full results can
be found in the Appendix in Tables~\ref{tab:collision_null} and
\ref{tab:collision_spacy}.}

To understand the findings better, Table~\ref{tab:counts} shows the number of
unique  \texttt{NORM}, \texttt{PREFIX}, \texttt{SUFFIX} and \texttt{SHAPE}
features found for each dataset.  First, note that having 2500 rows for the
\texttt{PREFIX} feature is too large, especially on datasets with Latin scripts.
In addition, the number of \texttt{SHAPE} rows is also below 500 for most 
datasets except WNUT 2017. This suggests that  the default, 2500, is more than
enough for many use cases.  In terms of collisions, we did not see significant
improvements from using multiple hash functions even if for example the CoNLL
datasets contain five times the number of the default 5000 rows for 
\texttt{NORM}. This is most likely due to the presence of other orthographic
features abating the possibility of collisions. 

However, we do see a different pattern for the larger OntoNotes dataset. Here,
we see a 5\% increase in F1-score from using three hash functions as opposed to
one, but we do not observe further improvements when using four. This is most
likely due to the much larger vocabulary size of OntoNotes compared to the rest
of the datasets. Based on these results, we will implement a parameter for
\texttt{MultiHashEmbed} to configure the number of hash functions.

\begin{table}[t]
    \centering
    \setlength{\tabcolsep}{3pt}
    \begin{tabular}{lcccc}
        \toprule
         & NORM & PREFIX & SUFFIX &  SHAPE \\
         \midrule
         CoNLL es & 26099 & 84 &  4369 & 314 \\
         CoNLL nl & 26036 & 85 & 3988 & 328\\
         WNUT2017 & 12837 & 92 & 5867 & 2103\\
         Archeology & 23112 & 139 & 4764 & 486\\
         AnEM & 7924 &  91 & 2359 & 184\\
         OntoNotes & 49397 & 110 & 7507 & 674 \\
         \bottomrule
    \end{tabular}
    \caption{Counts of the occurrence of orthographic features in the training sets.}
    \label{tab:counts}
\end{table}

\section{Discussion and Conclusion}

Word embeddings have a profound effect on the accuracy of modern NLP pipelines.
The choice of embedding architecture in spaCy is based on hash embeddings, where
we use the hashing trick to provide a memory-efficient alternative to 
traditional embeddings. In this report, we evaluated the effectiveness of
spaCy's \texttt{MultiHashEmbed}. We compared it with a traditional approach to
assess our architectural decisions and establish training and pipeline 
recommendations for our users. We found that using hash embeddings is
competitive with traditional embeddings in accuracy while being more
memory-efficient, and that our additional orthographic features can improve
performance.  These results support the basic architectural choices in spaCy. 

However, we also found some surprising results. The benefit of additional
orthographic features is subtle: on the AnEM dataset they improve the overall
performance, but at the same time degrade the F1-scores on entities that were 
seen during training. This finding indicates that although the defaults provide
good performance, fine-grained evaluation for each dataset is important.
Additionally, we were surprised to find that using more than one hash function
does not lead to performance gains in most of the datasets we considered for
this report. In the future we will implement a version of
\texttt{MultiHashEmbed} that parametrizes the number of hash functions. We
believe that this parameter can lead to significant improvements in speed and
power usage. 

Given these findings, we recommend spaCy users to:

\begin{itemize}
    \item \textbf{Thoroughly inspect the data before training an NER pipeline.}
    We provide a utility in spaCy that reports useful entity statistics and
    problems like invalid entity annotations or low data labels.\footnote{Read
    more about \texttt{debug data}: \url{https://spacy.io/api/cli\#debug-data}} 
    \item \textbf{Experiment with the use of orthographic features.} Based on
    one's dataset, certain features can be relevant or irrelevant. We recommend
    an investigation similar to Section \ref{sec:orth} to determine which
    features should be included.
    \item \textbf{Experiment with the number of rows in the lookup table and the
    number of hash functions.} We discovered that there are many ways to reduce
    the memory footprint of a spaCy pipeline without considerable degradation in
    performance. A dataset may benefit from a lower number of rows and a single
    hash function.
    \item \textbf{Make use of pretrained embeddings distributed by spaCy.} We
    provide high quality pretrained embeddings for multiple languages. They can
    significantly boost performance on smaller datasets.
\end{itemize}

The default configuration in spaCy can deliver good performance out of the box.
However, they are biased towards providing a good performance and efficiency
trade-off for the pretrained pipelines\footnote{\url{https://spacy.io/models}}
often trained on larger and more diverse corpora such as OntoNotes. 

As such, we encourage spaCy users to follow our recommendations and experiment
with various parameters potentially making their production pipelines more
efficient, with little to no cost in performance.

\newpage
\bibliographystyle{apalike}
\bibliography{main.bib}

\newpage
\appendix
\section{Appendix}

\subsection{Dataset statistics}

Table \ref{tab:dataset_statistics} shows the training set characteristics for
each dataset we used in the experiments whereas Table 
\ref{tab:seen_and_unseen_statistics} shows the number of unseen and seen
entities for each split.

\begin{table}[h]
    \centering
    \setlength{\tabcolsep}{2pt}
    \begin{tabular}{lcccccccc}
        \toprule
         & Documents & Tokens & Classes & Entities & Doc length &  Ent length & Vocab & Unknown \\
         \midrule
         CoNLL es &  8323 & 264715  & 4 & 2.25 & 31.8 & 1.74 & 26099 & 2819  \\
         CoNLL nl & 287 & 202644 & 4 & 46.5 & 706 & 1.4 & 27804 & 4148\\
         WNUT2017 & 3394 & 62730 & 6 & 0.58 & 18.5 & 1.6 & 14878 & 4911\\
         Archeology & 26804 & 351235 & 6 & 0.92 & 13.10 & 1.37 & 25404 & 8921 \\
         AnEM & 2252 & 57894 & 11 & 0.66 & 25.7 & 1.5 & 8711 & 1057  \\
         OntoNotes & 18127 & 2097259 & 18 & 6.95 & 115.69 & 1.86 & 57131 & 7310 \\
         \bottomrule
    \end{tabular}
    \caption{Training set characteristics.}
    \label{tab:dataset_statistics}
\end{table}

For Table \ref{tab:dataset_statistics}, the \textit{Entities} column shows the
number of entities per document.  \textit{Doc length} and \textit{Ent length}
indicate the mean number of tokens within documents and entity spans
respectively. The \textit{Vocab} column shows the size of the vocabulary,
estimated based on the raw orthographic form of tokens produced by spaCy's
tokenizer. The \textit{Unknown} column indicates the number of tokens that are
not in the vocabulary of the pretrained embeddings.

\begin{table}[h]
    \centering
    \begin{tabular}{lcccc}
        \toprule 
          & \multicolumn{2}{c}{Dev} & \multicolumn{2}{c}{Test} \\
          & Seen & Unseen & Seen & Unseen \\
         \midrule
         CoNLL es & 2300 & 2052 & 2214 & 1345\\
         CoNLL nl & 1039 & 1577 & 1797 & 2144\\
         WNUT2017 & 0 & 836 & 0 & 1079 \\
         Archeology & 2764 & 492 & 2669 & 482 \\
         AnEM & 290 & 110 & 548 & 708 \\
         OntoNotes & 13964 & 6004 & 8620 & 3517 \\
         \bottomrule
    \end{tabular}
    \caption{The total number of seen and unseen entities for the development and test sets for each corpus. Notice that WNUT 2017 only contains unseen entities.}
    \label{tab:seen_and_unseen_statistics}
\end{table}

\subsection{Minimum frequency}
\label{sec:minfreq}

Throughout our experiments, we used a minimum frequency of 10 in
\texttt{MultiEmbed}. We are interested if this choice has biased our comparison.
Hence, we compare \texttt{MultiHashEmbed} with default settings against 
\texttt{MultiEmbed} using minimum document frequencies of 10, 5, and 1. 

\begin{figure}[t]
    \centering
    \includegraphics[width=\textwidth]{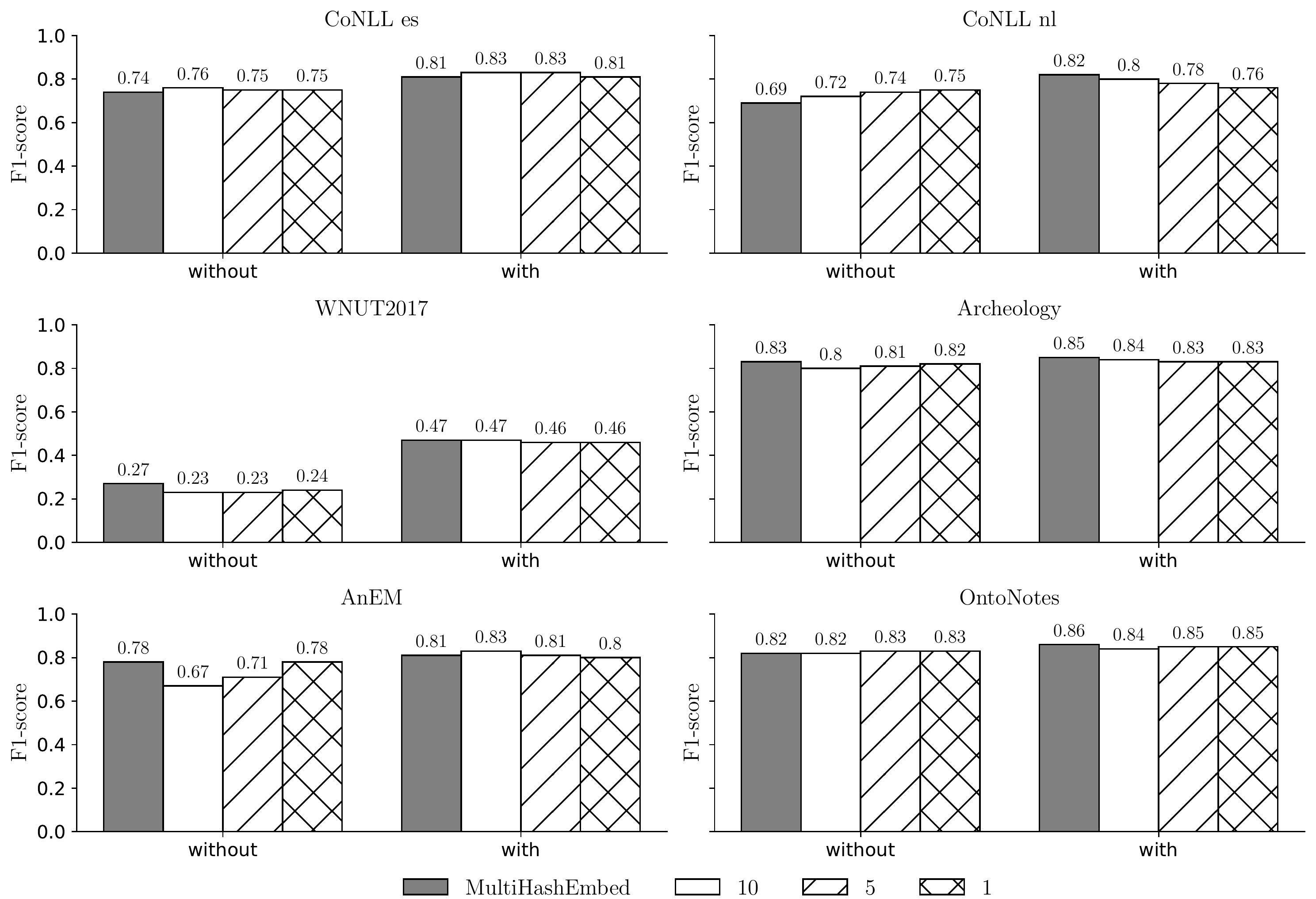}
    \caption{
        Characterizing \texttt{MultiEmbed} performance with and without
        pretrained embeddings across different minimum frequency values (i.e.,
        10, 5, and 1). Evaluated on the development set.
    }
    \label{fig:me_min_freq}
\end{figure}

We show the results with and without pretrained embeddings in Figure
\ref{fig:me_min_freq}.\footnote{Full results reported in 
Table~\ref{tab:me_min_freq_null} and \ref{tab:me_min_freq_spacy}.} We find that
\texttt{MultiEmbed} performs the best overall with minimum frequency of 10 when
used together with pretrained embeddings. 

However, the results are inconsistent when we test without static vectors. On
Spanish ConLL, Archeology, WNUT 2017, and OntoNotes, the performance is not
affected by the minimum frequency threshold\textemdash only a $\pm$2 change in
the F1-score.  We did not observe this in AnEM: changing the minimum frequency
cutoff from 10 to 1 caused a dramatic increase of 11 points.  Something similar
happened in Dutch ConLL, reducing the minimum frequency to 1 caused a 6 point
increase compared to \texttt{MultiHashEmbed}.

When taking these results into consideration, it seems that \texttt{MultiEmbed}
has the potential to outperform \texttt{MultiHashEmbed} (using default settings)
on ConLL, while allowing its lookup table to only have up to 27k rows for the
\texttt{NORM} orthographic feature.

With that said, our choice of 10 as the minimum document frequency cutoff for
\texttt{MultiEmbed} still gives the most consistent results. Hence we maintained
the same value for all benchmarking scenarios.

\begin{figure}[t]
    \centering
    \includegraphics[width=\textwidth]{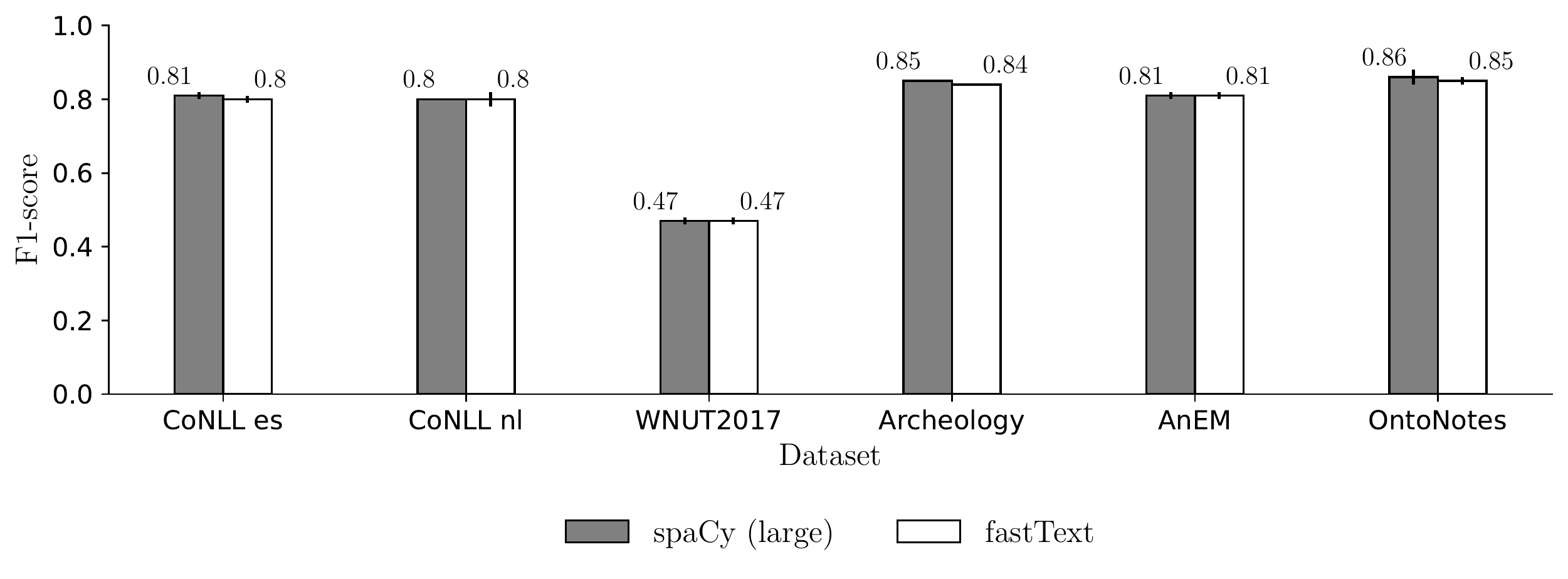}
    \caption{Comparing large spaCy vectors (3.4.3) against fastText vectors.}
    \label{fig:spacy_v_fasttext}
\end{figure}

\subsection{Pretrained embeddings distributed by spaCy vs. fastText}
\label{sec:spacyfasttext}

The fastText vectors are another set of commonly-used word embeddings outside
the spaCy ecosystem. They are available for multiple
languages.\footnote{\url{https://fasttext.cc/}} In Figure
\ref{fig:spacy_v_fasttext}, we compare the performance between spaCy (large) and
fastText vectors when used alongside \texttt{MultiHashEmbed}.  We find that
there is almost no difference in performance. These results indicate that the
vectors distributed with spaCy provide similar performance to fastText
vectors.\footnote{The complete results are presented in
Table~\ref{tab:mhe_fasttext_v_spacy}.}

\section{Tables with complete results}

\begin{table}[H]
    \centering
    \begin{tabular}{c|ccc|ccc}
        \toprule
         & \multicolumn{3}{c|}{\texttt{MultiHashEmbed}} & \multicolumn{3}{c}{\texttt{MultiEmbed}}                   \\
                        & Precision           & Recall           & F1-score         & Precision             & Recall             & F1-score             \\
        \midrule
        CoNLL es        & 0.83$\pm$0.01 & 0.82$\pm$0.01 & 0.82$\pm$0.01 & 0.84$\pm$0.01 & 0.84$\pm$0.00 & 0.84$\pm$0.00  \\
        CoNLL nl        & 0.83$\pm$0.00 & 0.82$\pm$0.01 & 0.83$\pm$0.00 & 0.84$\pm$0.01 & 0.83$\pm$0.01 & 0.83$\pm$0.01  \\
        WNUT2017        & 0.53$\pm$0.02 & 0.25$\pm$0.04 & 0.34$\pm$0.03 & 0.53$\pm$0.03 & 0.26$\pm$0.02 & 0.35$\pm$0.02  \\
        Archeology      & 0.83$\pm$0.01 & 0.82$\pm$0.01 & 0.83$\pm$0.01 & 0.83$\pm$0.01 & 0.82$\pm$0.01 & 0.83$\pm$0.00  \\
        AnEM            & 0.66$\pm$0.01 & 0.57$\pm$0.02 & 0.61$\pm$0.01 & 0.65$\pm$0.02 & 0.58$\pm$0.02 & 0.61$\pm$0.01  \\
        OntoNotes       & 0.84$\pm$0.02 & 0.83$\pm$0.03 & 0.84 $\pm$0.01 & 0.83$\pm$0.04 & 0.82 $\pm$ 0.01 & 0.83 $\pm$ 0.02 \\
        \bottomrule
    \end{tabular}
    \caption{Comparison between \texttt{MultiHashEmbed} with default number of rows and \texttt{MultiEmbed} using pretrained embeddings distributed by spaCy. Results are reported on the test set.}
    \label{tab:mhe_v_me_default_spacy_test}
\end{table}

\begin{table}[h]
    \centering
    \begin{tabular}{c|ccc|ccc}
        \toprule
         & \multicolumn{3}{c|}{\texttt{MultiHashEmbed}} & \multicolumn{3}{c}{\texttt{MultiEmbed}}                   \\
                        & Precision           & Recall           & F1-score         & Precision             & Recall             & F1-score             \\
        \midrule
        CoNLL es        & 0.84$\pm$0.01 & 0.86$\pm$0.03 & 0.85$\pm$0.01 & 0.84$\pm$0.01 & 0.84$\pm$0.00 & 0.84$\pm$0.00  \\
        CoNLL nl        & 0.84$\pm$0.01 & 0.84$\pm$0.01 & 0.84$\pm$0.02 & 0.84$\pm$0.01 & 0.83$\pm$0.01 & 0.83$\pm$0.01  \\
        WNUT2017        & 0.56$\pm$0.01 & 0.27$\pm$0.01 & 0.36$\pm$0.02 & 0.53$\pm$0.03 & 0.26$\pm$0.02 & 0.35$\pm$0.02  \\
        Archeology      & 0.84$\pm$0.01 & 0.83$\pm$0.01 & 0.83$\pm$0.01 & 0.83$\pm$0.01 & 0.82$\pm$0.01 & 0.83$\pm$0.00  \\
        AnEM            & 0.67$\pm$0.01 & 0.57$\pm$0.02 & 0.62$\pm$0.01 & 0.65$\pm$0.02 & 0.58$\pm$0.02 & 0.61$\pm$0.01  \\
        OntoNotes       & 0.85$\pm$0.01 & 0.84$\pm$0.03 & 0.85$\pm$0.02 & 0.83$\pm$0.04 & 0.82$\pm$0.01 & 0.83$\pm$0.02 \\
        \bottomrule
    \end{tabular}
    \caption{Comparison between \texttt{MultiHashEmbed} with adjusted number of rows and \texttt{MultiEmbed} using pretrained embeddings distributed by spaCy. Results are reported on the test set. Adjusted rows for \texttt{MultiHashEmbed}.}
    \label{tab:mhe_v_me_adjusted_rows_spacy_test}
\end{table}

\begin{table}[H]
    \centering
    \begin{tabular}{c|ccc|ccc}
        \toprule
         & \multicolumn{3}{c|}{\texttt{MultiHashEmbed}} & \multicolumn{3}{c}{\texttt{MultiEmbed}}                   \\
                        & Precision           & Recall           & F1-score          & Precision             & Recall             & F1-score             \\
        \midrule
        CoNLL es        & 0.77$\pm$0.01 & 0.77$\pm$0.01 & 0.77$\pm$0.00 & 0.77$\pm$0.01 & 0.77$\pm$0.01 & 0.77$\pm$0.01  \\
        CoNLL nl        & 0.75$\pm$0.01 & 0.72$\pm$0.01 & 0.73$\pm$0.01 & 0.77$\pm$0.00 & 0.74$\pm$0.01 & 0.75$\pm$0.00  \\
        WNUT2017        & 0.27$\pm$0.03 & 0.12$\pm$0.02 & 0.17$\pm$0.02 & 0.29$\pm$0.00 & 0.14$\pm$0.01 & 0.19$\pm$0.01  \\
        Archeology      & 0.83$\pm$0.01 & 0.81$\pm$0.00 & 0.82$\pm$0.00 & 0.82$\pm$0.00 & 0.78$\pm$0.01 & 0.80$\pm$0.00  \\
        AnEM            & 0.60$\pm$0.01 & 0.49$\pm$0.01 & 0.54$\pm$0.01 & 0.56$\pm$0.02 & 0.36$\pm$0.03 & 0.44$\pm$0.01  \\
        OntoNotes       & 0.82$\pm$0.01 & 0.79$\pm$0.01 & 0.81$\pm$0.02 & 0.81$\pm$0.02 & 0.81$\pm$0.01 & 0.81$\pm$0.01 \\
        \bottomrule
    \end{tabular}
    \caption{Comparison between \texttt{MultiHashEmbed} with default number of rows and \texttt{MultiEmbed} without pretrained embeddings distributed by spaCy. Results are reported on the test set.}
    \label{tab:mhe_v_me_default_null_test}
\end{table}

\begin{table}[h]
    \centering
    \begin{tabular}{c|ccc|ccc}
        \toprule
         & \multicolumn{3}{c|}{\texttt{MultiHashEmbed}} & \multicolumn{3}{c}{\texttt{MultiEmbed}}                   \\
                        & Precision           & Recall           & F1-score          & Precision             & Recall             & F1-score             \\
         \midrule
        CoNLL es        & 0.78$\pm$0.01 & 0.77$\pm$0.02 & 0.78$\pm$0.02 & 0.77$\pm$0.01 & 0.77$\pm$0.01 & 0.77$\pm$0.01  \\
        CoNLL nl        & 0.77$\pm$0.02 & 0.73$\pm$0.00 & 0.75$\pm$0.03 & 0.77$\pm$0.00 & 0.74$\pm$0.01 & 0.75$\pm$0.00  \\
        WNUT2017        & 0.28$\pm$0.03 & 0.15$\pm$0.02 & 0.19$\pm$0.01 & 0.29$\pm$0.00 & 0.14$\pm$0.01 & 0.19$\pm$0.01  \\
        Archeology      & 0.84$\pm$0.01 & 0.80$\pm$0.00 & 0.81$\pm$0.01 & 0.82$\pm$0.00 & 0.78$\pm$0.01 & 0.80$\pm$0.00  \\
        AnEM            & 0.62$\pm$0.01 & 0.51$\pm$0.01 & 0.56$\pm$0.01 & 0.56$\pm$0.02 & 0.36$\pm$0.03 & 0.44$\pm$0.01  \\
        OntoNotes       & 0.82$\pm$0.02 & 0.82$\pm$0.01 & 0.82$\pm$0.01 & 0.81$\pm$0.02 & 0.81$\pm$0.01 & 0.81$\pm$0.01 \\
        \bottomrule
    \end{tabular}
    \caption{Comparison between \texttt{MultiHashEmbed} with adjusted number of rows and \texttt{MultiEmbed} without pretrained embeddings distributed by spaCy. Results are reported on the test set.}
    \label{tab:mhe_v_me_defaults_null_test}
\end{table}

\begin{table}[H]
    \centering
    \begin{tabular}{c|ccc|ccc}
    \toprule
         & \multicolumn{3}{c|}{\texttt{MultiHashEmbed}} & \multicolumn{3}{c}{\texttt{MultiEmbed}}                   \\
                        & Precision           & Recall           & F1-score         & Precision             & Recall             & F1-score             \\
        
        \midrule
        CoNLL es        & 0.59$\pm$0.02 & 0.60$\pm$0.01 & 0.60$\pm$0.01 & 0.64$\pm$0.02 & 0.64$\pm$0.02 & 0.65$\pm$0.01  \\
        CoNLL nl        & 0.61$\pm$0.02 & 0.57$\pm$0.00 & 0.59$\pm$0.02 & 0.63$\pm$0.03 & 0.58$\pm$0.00 & 0.59$\pm$0.01  \\
        WNUT2017        & 0.21$\pm$0.04 & 0.10$\pm$0.01 & 0.14$\pm$0.01 & 0.23$\pm$0.02 & 0.13$\pm$0.01 & 0.17$\pm$0.03  \\
        Archeology      & 0.33$\pm$0.01 & 0.38$\pm$0.00 & 0.35$\pm$0.03 & 0.30$\pm$0.01 & 0.42$\pm$0.04 & 0.35$\pm$0.02  \\
        AnEM            & 0.26$\pm$0.03 & 0.18$\pm$0.02 & 0.21$\pm$0.02 & 0.32$\pm$0.02 & 0.18$\pm$0.03 & 0.23$\pm$0.03  \\
        OntoNotes       & 0.54$\pm$0.02 & 0.58$\pm$0.01 & 0.56$\pm$0.01 & 0.62$\pm$0.01 & 0.60$\pm$0.02 & 0.61$\pm$0.01  \\
        \bottomrule
    \end{tabular}
    \caption{Comparison between \texttt{MultiHashEmbed} with default number of rows and \texttt{MultiEmbed} without pretrained embeddings distributed by spaCy. Results are reported on the unseen test set.}
    \label{tab:mhe_v_me_unseen_default_null_test}
\end{table}

\begin{table}[H]
    \centering
    \begin{tabular}{c|ccc|ccc}
        \toprule
         & \multicolumn{3}{c|}{\texttt{MultiHashEmbed}} & \multicolumn{3}{c}{\texttt{MultiEmbed}}                   \\
                        & Precision           & Recall           & F1-score         & Precision             & Recall             & F1-score             \\
        \midrule
        CoNLL es        & 0.74$\pm$0.01 & 0.74$\pm$0.02 & 0.74$\pm$0.01 & 0.74$\pm$0.02 & 0.73$\pm$0.03 & 0.73$\pm$0.02  \\
        CoNLL nl        & 0.72$\pm$0.03 & 0.69$\pm$0.01 & 0.70$\pm$0.02 & 0.74$\pm$0.03 & 0.72$\pm$0.01 & 0.73$\pm$0.03  \\
        WNUT2017        & 0.45$\pm$0.01 & 0.28$\pm$0.01 & 0.34$\pm$0.00 & 0.47$\pm$0.01 & 0.24$\pm$0.03 & 0.32$\pm$0.03  \\
        Archeology      & 0.37$\pm$0.02 & 0.42$\pm$0.01 & 0.40$\pm$0.02 & 0.37$\pm$0.02 & 0.50$\pm$0.02 & 0.43$\pm$0.01  \\
        AnEM            & 0.42$\pm$0.01 & 0.30$\pm$0.02 & 0.35$\pm$0.03 & 0.36$\pm$0.01 & 0.28$\pm$0.02 & 0.32$\pm$0.01  \\
        OntoNotes       & 0.64$\pm$0.02 & 0.65$\pm$0.02 & 0.65$\pm$0.01 & 0.64$\pm$0.02 & 0.69$\pm$0.01 & 0.66$\pm$0.02 \\
        \bottomrule
    \end{tabular}
    \caption{Comparison between \texttt{MultiHashEmbed} with default number of rows and \texttt{MultiEmbed} with pretrained embeddings distributed by spaCy. Results are reported on the unseen test set.}
    \label{tab:mhe_v_me_unseen_default_spacy_test}
\end{table}

\begin{table}[h]
    \centering
    \begin{tabular}{ccccc}
        \toprule
         & \multicolumn{4}{c}{Number of hash functions}\\
         &  1 & 2& 3 & 4 \\
        \midrule
        CoNLL es &  0.75$\pm$0.03 & 0.74$\pm$0.02 & 0.73$\pm$0.02 & 0.74$\pm$0.01 \\
        CoNLL nl & 0.70$\pm$0.01 & 0.68$\pm$0.02 & 0.70$\pm$0.02 & 0.69$\pm$0.01  \\
        WNUT2017 & 0.22$\pm$0.03 & 0.22$\pm$0.02 & 0.22$\pm$0.01 & 0.23$\pm$0.02  \\
        Archeology &  0.82$\pm$0.02 & 0.82$\pm$0.01 & 0.82$\pm$0.02 & 0.82$\pm$0.00 \\
        AnEM & 0.77$\pm$0.03 & 0.77$\pm$0.03 & 0.78$\pm$0.02 & 0.78$\pm$0.01  \\   
        OntoNotes & 0.77$\pm$0.03 & 0.79 $\pm$0.01 & 0.82 $\pm$0.01 & 0.82 $\pm$0.02 \\ 
        \bottomrule
    \end{tabular}
   \caption{The effect of changing the number of seeds for \texttt{MultiHashEmbed} on dev set performance (F1-score). Without pretrained embeddings.}
    \label{tab:collision_null}
\end{table}

\begin{table}[h]
    \centering
    \begin{tabular}{ccccc}
        \toprule
         & \multicolumn{4}{c}{Number of hash functions}\\
         &  1 & 2& 3 & 4 \\
        \midrule
        CoNLL es & 0.81$\pm$0.01 & 0.81$\pm$0.02 & 0.81$\pm$0.02 & 0.81$\pm$0.00 \\
        CoNLL nl &  0.80$\pm$0.02 & 0.80$\pm$0.02 & 0.81$\pm$0.01 & 0.80$\pm$0.00 \\
        WNUT2017 & 0.47$\pm$0.03 & 0.47$\pm$0.03 & 0.47$\pm$0.01 & 0.47$\pm$0.01 \\
        Archeology &  0.83$\pm$0.00 & 0.83$\pm$0.01 & 0.84$\pm$0.02 & 0.85$\pm$0.00 \\
        AnEM & 0.80$\pm$0.02 & 0.80$\pm$0.02 & 0.80$\pm$0.01 & 0.81$\pm$0.01 \\    
        OntoNotes & 0.78$\pm$0.03 & 0.81 $\pm$0.01 & 0.83 $\pm$0.01 & 0.83 $\pm$0.01 \\ 
        \bottomrule
    \end{tabular}
   \caption{The effect of changing the number of seeds for \texttt{MultiHashEmbed} on dev set performance (F1-score). Using spaCy vectors.}
    \label{tab:collision_spacy}
\end{table}

\begin{table}[h]
    \centering
    \begin{tabular}{ccccc}
        \toprule
         &  \texttt{MHE} & \texttt{ME} min-freq=10 & \texttt{ME} min-freq=5 & \texttt{ME} min-freq=1 \\
         \midrule
        CoNLL es &          0.74 & 0.76 & 0.75 & 0.75 \\
        CoNLL nl &          0.69 & 0.72 & 0.74 & 0.75 \\
        WNUT2017 &          0.27 & 0.23 & 0.23 & 0.24 \\
        Archeology  &   0.83 & 0.80 & 0.81 & 0.82 \\
        AnEM  &             0.78 & 0.67 & 0.71 & 0.78 \\    
        OntoNotes &         0.82 & 0.82 & 0.83 & 0.83  \\
        \bottomrule
    \end{tabular}
    \caption{Results reported are on F1-scores on the dev sets. \texttt{MHE} stands for 
    \texttt{MultiHashEmbed} and \texttt{ME} for \texttt{MultiEmbed}. Without pretrained embeddings.}
    \label{tab:me_min_freq_null}
\end{table}

\begin{table}[h!]
    \centering
    \begin{tabular}{ccccc}
        \toprule
         &  \texttt{MHE} & \texttt{ME} min-freq=10 & \texttt{ME} min-freq=5 & \texttt{ME} min-freq=1 \\
         \midrule
        CoNLL es &          0.81 & 0.83 & 0.83 & 0.81 \\
        CoNLL nl &          0.82 & 0.80 & 0.78 & 0.76 \\
        WNUT2017 &          0.47 & 0.47 & 0.46 & 0.46 \\
        Archeology &   0.85 & 0.84 & 0.83 & 0.83 \\
        AnEM  &             0.81 & 0.83 & 0.81 & 0.80 \\    
        OntoNotes &         0.86 & 0.84 & 0.85 & 0.85 \\
        \bottomrule
    \end{tabular}
    \caption{Results reported are on F1-scores on the dev sets. \texttt{MHE} stands for 
    \texttt{MultiHashEmbed} and \texttt{ME} for \texttt{MultiEmbed}.  Using pretrained embeddings distributed by spaCy.}
    \label{tab:me_min_freq_spacy}
\end{table}

\begin{table}[h!]
    \centering
    \begin{tabular}{c|ccc|ccc}
        \toprule
         & \multicolumn{3}{c|}{fastText} & \multicolumn{3}{c}{spaCy (large)}                   \\
                        & Precision      & Recall           & F1-score          & Precision      & Recall         & F1-score          \\
        \midrule
        CoNLL es        & 0.80$\pm$0.01  & 0.80$\pm$0.00    & 0.80$\pm$0.01    & 0.82$\pm$0.01  & 0.80$\pm$0.01  & 0.81$\pm$0.01    \\
        CoNLL nl        & 0.81$\pm$0.01  & 0.80$\pm$0.01    & 0.80$\pm$0.02    & 0.80$\pm$0.00  & 0.80$\pm$0.01  & 0.80$\pm$0.00    \\
        WNUT2017        & 0.62$\pm$0.02  & 0.37$\pm$0.01    & 0.47$\pm$0.01    & 0.67$\pm$0.06  & 0.36$\pm$0.01  & 0.47$\pm$0.01    \\
        Archeology  & 0.84$\pm$0.01  & 0.85$\pm$0.01    & 0.84$\pm$0.00    & 0.85$\pm$0.01  & 0.84$\pm$0.02  & 0.85$\pm$0.00    \\
        AnEM            & 0.82$\pm$0.02  & 0.80$\pm$0.01    & 0.81$\pm$0.01    & 0.83$\pm$0.10  & 0.79$\pm$0.01  & 0.81$\pm$0.01    \\
        OntoNotes       & 0.84$\pm$0.02 & 0.86$\pm$0.02 & 0.85 $\pm$0.01 & 0.85$\pm$0.02 & 0.85$\pm$0.01 & 0.86$\pm$0.02 \\
        \bottomrule
    \end{tabular}
    \caption{Comparison between fastText vectors and pretrained embeddings distributed by spaCy. for \texttt{MultiHashEmbed}. Default number of rows.}
    \label{tab:mhe_fasttext_v_spacy}
\end{table}

\section{Configuration}
\label{sec:configuration}
This section shows the partial configuration files we used throughout the
experiments.\footnote{For more information about spaCy configuration files,
visit the documentation: \url{https://spacy.io/usage/training\#config}} To make
things more succinct, we removed unnecessary parameters such as path names and
pretraining arguments.  Listing \ref{listing:mhe} shows the configuration file
for \texttt{MultiHashEmbed}.

\begin{lstlisting}[
    frame=shadowbox,
    caption=Partial configuration file for \texttt{MultiHashEmbed},
    language=Ini,
    numbers=left,
    label={listing:mhe},
]
[components]

[components.ner.model]
@architectures = "spacy.TransitionBasedParser.v2"
state_type = "ner"
extra_state_tokens = false
hidden_width = 64
maxout_pieces = 2
use_upper = true
nO = null

[components.ner.model.tok2vec]
@architectures = "spacy.Tok2VecListener.v1"
width = ${components.tok2vec.model.encode.width}
upstream = "*"

[components.tok2vec]
factory = "tok2vec"

[components.tok2vec.model]
@architectures = "spacy.Tok2Vec.v2"

[components.tok2vec.model.embed]
@architectures = "spacy.MultiHashEmbed.v2"
width = ${components.tok2vec.model.encode.width}
attrs = ["NORM","PREFIX","SUFFIX","SHAPE"]
rows = [5000,2500,2500,2500]
include_static_vectors = true

[components.tok2vec.model.encode]
@architectures = "spacy.MaxoutWindowEncoder.v2"
width = 256
depth = 8
window_size = 1
maxout_pieces = 3

[training]
dropout = 0.1
accumulate_gradient = 1
patience = 1600
max_epochs = 0
max_steps = 20000
eval_frequency = 200

[training.optimizer]
@optimizers = "Adam.v1"
beta1 = 0.9
beta2 = 0.999
L2_is_weight_decay = true
L2 = 0.01
grad_clip = 1.0
use_averages = false
eps = 0.00000001
learn_rate = 0.001
\end{lstlisting}

The parameter \texttt{include\_static\_vectors} refers to whether pretrained
embeddings will be used in the spaCy pipeline. This setting is relevant in our
experiments in Section \ref{sec:mhe_v_me} and \ref{sec:minfreq} where we
benchmarked performance with and without pretrained embeddings.

We also implemented \texttt{MultiEmbed} that is similar to
\texttt{MultiHashEmbed} except that it uses a regular lookup instead of the
hashing trick. Here, we only need to change the 
\texttt{components.tok2vec.model.embed} section as seen in Listing
\ref{listing:me}:

\begin{lstlisting}[
    frame=shadowbox,
    caption=Partial configuration file for \texttt{MultiEmbed},
    language=Ini,
    numbers=left,
    label={listing:me},
]
[components.tok2vec.model.embed]
@architectures = "spacy.MultiEmbed.v1"
attrs = ["NORM","PREFIX","SUFFIX","SHAPE"]
width = ${components.tok2vec.model.encode.width}
include_static_vectors = true
unk = 0
\end{lstlisting}

Lastly, in order to perform the experiments found in Section
\ref{sec:hash_func}, we modified \texttt{MultiHashEmbed} to accept a variable
number of hash functions. We call this \texttt{MultiFewerHashEmbed}. The
configuration is again similar to \texttt{MultiHashEmbed} with only a change to
the \texttt{components.tok2vec.model.embed} parameter:

\begin{lstlisting}[
    frame=shadowbox,
    caption=Partial configuration file for \texttt{MultiFewerHashEmbed},
    language=Ini,
    numbers=left,
    label={listing:mfhe},
]
[components.tok2vec.model.embed]
@architectures = "spacy.MultiFewerHashEmbed.v1"
num_hashes = 1
width = ${components.tok2vec.model.encode.width}
attrs = ["NORM","PREFIX","SUFFIX","SHAPE"]
rows = [5000,2500,2500,2500]
include_static_vectors = true
\end{lstlisting}

\end{document}